\def\year{2017}\relax
\documentclass[letterpaper]{article}
\usepackage{aaai17}
\usepackage{times}
\usepackage{helvet}
\usepackage{courier}
\usepackage{amsmath}
\usepackage{bm}
\usepackage{graphicx}
\usepackage{mathrsfs}
\usepackage{balance}

\DeclareMathOperator*{\argmax}{argmax}
\DeclareMathOperator{\softmax}{softmax}
\DeclareMathOperator{\f}{f}
\DeclareMathOperator{\g}{g}

\DeclareMathOperator{\LSTM}{LSTM}

\makeatletter
\def\blfootnote{\gdef\@thefnmark{}\@footnotetext}
\makeatother

\begin{document}
	
	\title{Neural Programming by Example}
	\author{Chengxun Shu\\
%		School of Electronics and Information Engineering\\
		Beihang University \\
        Beijing 100191, China\\
		shuchengxun@163.com
		\And
		Hongyu Zhang\\
 %       School of Electrical Engineering and Computer Science\\
		The University of Newcastle \\
        Callaghan, NSW 2308, Australia\\
		hongyu.zhang@newcastle.edu.au
	}
	\maketitle
	\begin{abstract}
		Programming by Example (PBE) targets at automatically inferring a computer program for accomplishing a certain task from sample input and output. In this paper, we propose a deep neural networks (DNN) based PBE model called Neural Programming by Example (NPBE), which can learn from input-output strings and induce programs that solve the string manipulation problems. 
		%NPBE has four components: a string encoder, an input-output analyzer, a program generator and a symbol selector. String encoder can encode the input and output strings to some summary embeddings and also a list of character level embeddings for later attention mechanism. Input-output analyzer is fed with the input and output string summary embeddings to generate the transformation embedding describing the relationship between input and output strings. Program generator can decode the transformation embedding and generate program embeddings which can refer to the output of previous program embeddings, thus the entire framework can learn to compose complicated programs. Symbol selector use the output of program generator to generate human readable program symbols. 
% 		For 74.6\% input-output pairs in our test dataset, NPBE successfully generates the corresponding program. \todo{first, precisely it's not 74.6\% input-output pairs, but 74.6\% programs. second, I'm not sure whether we should report this number here. why not report top5 accuracy? if report 74.6\%, i think we should state that it's the result when we only gives one prediction}
% 		The results show that our model is effective and is a step towards teaching DNN to generate computer programs.
		Our NPBE model has four neural network based components: a string encoder, an input-output analyzer, a program generator, and a symbol selector. We demonstrate the effectiveness of NPBE by training it end-to-end to solve some common string manipulation problems in spreadsheet systems. The results show that our model can induce string manipulation programs effectively. Our work is one step towards teaching DNN to generate computer programs.
		
	\end{abstract}
	
	\section{Introduction}
	Programming by Example (PBE, also called programming by demonstration, or inductive synthesis) \cite{lieberman2001your,cypher1993watch,gulwani2011automating} gives machines the ability to reason and generate new programs without substantial amount of human supervision. In PBE systems, users (often non-professional programmers) provide a machine with input-output examples of a task they would like to perform and the machine automatically infers a program to accomplish the task. The concept of PBE has been successfully used for string manipulation in spreadsheet systems such as Microsoft Excel \cite{gulwani2015inductive}, computer-aided education \cite{gulwani2014example} and data extracting systems \cite{le2014flashextract}.
	
	%PBE is challenging because users usually give only a few examples for a task.
	As an example, if a user provides the following input and output examples:
	\begin{center}
		\emph{\space john@example.com $\Rightarrow$ john}\\		
		\emph{\space james@company.com $\Rightarrow$ james}\\
	\end{center}
	A PBE system should understand that the user would like to extract the user name from the email address. It will automatically synthesize a program \textit{Select(Split(x, ‘@’), 0)}, where $x$ is the input string, $Split$ is to split a string according to a delimiter, and $Select$ is to select a substring from an array of strings. Given a new email address \emph{jacob@test.com}, the program will output the string \emph{jacob}.
	
	%Existing PBE solutions mainly adopt a kind of search technique to find a composition of predefined functions (such as string split and concatenation) that satisfy the input/output examples. %Although effective, the existing solutions could require more time to search when the search space is increasing. Furthermore, most existing solutions require end users to enter accurate input and output examples, without the ability to handle possible noises in examples. 
	
	%It is challenging to train a computer to learn program. 
	Lau et al. \shortcite{lau2003programming} applied version space algebra to search for possible programs. More recent PBE methods (Gulwani 2011) mainly adopt search technique to find a composition of predefined functions (such as string Split and Concatenation) that satisfies the input-output examples. These methods create a Directed Acyclic Graph (DAG) and search through the sequences of functions that can generate the output string from a given input state. These methods can generate complex string manipulation programs effectively, but require the design of complex program synthesis algorithms. 
	
	In this paper, we propose a Deep Neural Networks (DNN) based approach to Programming by Example. We train neural networks to automatically infer programs from input-output examples. During the past few years, research on DNN has achieved significant results in a variety of fields such as computer vision \cite{krizhevsky2012imagenet}, speech recognition \cite{mohamed2012acoustic}, natural language processing \cite{collobert2011natural}, and API learning \cite{Gu2016}. Recently, researchers have explored the feasibility of applying DNN to solve programming and computation related problems \cite{neelakantan2015neural,reed2015neural,graves2014neural,kurach2015neural}. Our work is based on the similar idea of applying DNN to infer and execute computer programs. Different from the existing work, we target at the PBE problem and train a neural PBE model with triples of input, output and program.
	
	Our approach, called NPBE (Neural Programming by Example), teaches DNN to compose a set of predefined atomic operations for string manipulation. Given an input-output string pair, the NPBE model is trained to synthesize a program - a sequence of functions and corresponding arguments that transform the input string to the output string. The program is generated from the atomic functions and one function may use the execution results of previous functions. Thus the model is able to compose complex programs using only several predefined operations.
	%Such a program is useful for transforming new input strings in a spreadsheet system.
	%The arguments of each function are also automatically recommended. %The atomic functions can be easily extended to strengthen the system's ability. %The execution of NPBE is totally symbolic so it's fast and extensible.%
	
	We have experimentally evaluated NPBE on a large number of input-output strings for 45 string manipulation tasks and the results are encouraging. We find that our model can generalize beyond the training input/output strings and the training argument settings. Our work is one of the early attempts to apply DNN to the problem of Programming by Example.
   %Our work is an early attempt to apply DNN to the Programming by Example problems concerning string manipulations.
	
	\section{Related Work}
	%Machine learning methods have been applied to solve PBE problems. For example, Menon et al. \shortcite{menon2013machine} propose a machine learning framework to reduce the search space of possible programs. Lau et al. \shortcite{lau2003programming} apply version space algebra to efficiently search possible programs. Given input-output pairs, genetic programming \cite{banzhaf1998genetic} can evolve useful programs from candidate populations. Liang et al. \shortcite{liang2010learning} propose a hierarchical Bayesian approach to learn simple programs given only a few examples.
	
	Reed et al. \shortcite{reed2015neural} developed a framework called Neural Programmer-Interpreter to induce and execute programs using neural networks. Their method treats programs as embeddings and uses neural networks to generate functions and arguments for program execution. Their model is trained with the supervision of execution traces and can be used in many different scenarios. However, their model cannot be directly applied to the problem of PBE as the input to their model is the environment encoded with the task, while our work is dedicated to PBE and the input to our model are input-output examples.   
    %One main difference between their work and ours is that the input to their model is the environment encoded with the task while the input to our model is the input-output examples. 
    Neural Programmer \cite{neelakantan2015neural} is a neural network augmented with a set of operations that can be called over several steps. It is trained to output the result of program execution, while our model is trained to output the program represented by symbols. Neural Enquirer \cite{yin2015neural} is a fully neural, end-to-end differentiable model capable of modeling and executing table query related programs. The execution of Neural Enquirer is ``softly" on data table using neural networks, while our work does not apply soft execution to input-output strings. %Riedel et al. \shortcite{Riedel2016Programming} proposed a differentiable abstract machine for the language Forth which use only program sketches and input-output pairs as input. It can handle more complex programs but needs sketches inputted.
{Bo{\v s}njak} et al. \shortcite{2016arXiv160506640B} proposed a neural implementation of an abstract machine for the language Forth. It can learn program behaviour trained from input-output data. However, it requires program sketches as input.   
	
	%Neural Programmer uses a neural network model to select operations during query processing. While the query planning (i.e., which operation to execute at each time step) phase is modeled softly using neural networks
	
	%But both of them execute the actual operations softly on the data which can be a huge burden when operations or data are huge-amount. Our work only executes an atom function symbolically and assumes further operations know what have been done. Compared to soft execution, our method is faster but maybe less accurate. 
	%Our model is also related to the work that uses recurrent neural networks to solve difficult problems.
	Our model is also related to the work that uses recurrent neural networks to solve programming and computation related problems. Graves et al. \shortcite{graves2014neural} developed Neural Turing Machine which is capable of learning and executing simple programs using an external memory. Zaremba et al. \shortcite{zaremba2015learning} used execution traces to train recurrent neural networks to learn simple algorithms. Ling et al. \shortcite{ling2016latent} developed a model to generate program code from natural language and structured specification. Pointer Networks \cite{vinyals2015pointer} uses an attentional recurrent model to solve difficult algorithmic problems. 
	
	Some machine learning methods were also proposed to tackle PBE problems. Lau et al. \shortcite{lau2003programming} applied version space algebra to efficiently search for possible programs. Given input-output pairs, genetic programming \cite{banzhaf1998genetic} can evolve useful programs from candidate populations. Liang et al. \shortcite{liang2010learning} proposed a hierarchical Bayesian approach to learn simple programs given only a few examples. Melon et al. \shortcite{menon2013machine} used machine learning to speed up the searching for possible programs by learning weights related to textual features.  Their method needs carefully designed features to reduce search space, while our method reduces search space and avoids feature engineering through learning representations using DNN. 
    %Their method needs carefully designed features to reduce search space, while our method uses DNN to learn representations thus reduces search space and avoids feature engineering.
    %Compared to their method which needs careful designed features to reduce search space, our method uses DNN to learn representations which can reduce search space and avoids feature engineering.
	
	\section{The NPBE Model}
	%Even though our NPBE model is not restricted to strings, we demonstrate it by using the task of string manipulation program induction. 
	
	\emph{Problem Statement}: Let $S$ denote the set of strings. For an input string $x\in S$ and an output string $y\in S$, our model is to induce a function $\f\in S^S$, so that $y=\f(x)$. The input to our model is the input-output pair $z:=(x,y)\in S^2$, and we wish to get the correct function $\f$, so when the user input another string $\bar{x}\in S$, we can generate the desired output $\bar{y}=\f(\bar{x})$ without the need of user specifying the function explicitly.
	
	\begin{figure}[ht]
		\centering
		\includegraphics[width=1\linewidth]{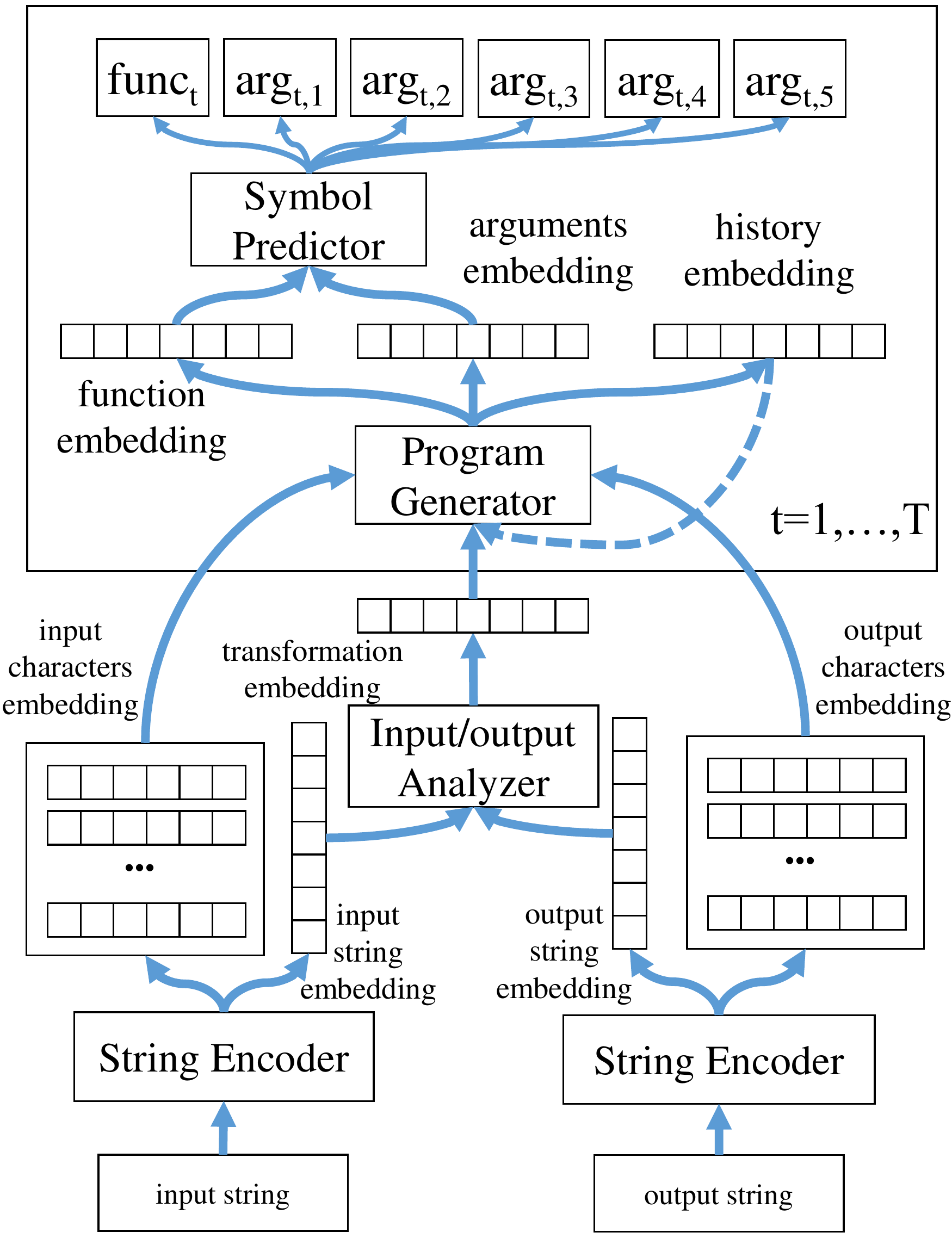}
		\caption{Overall architecture of NPBE.}
		\label{fig:overall}
	\end{figure}
	
	The proposed NPBE model (Figure \ref{fig:overall}) consists of four modules:
	\begin{itemize}
		\item A string encoder to encode input and output strings;
		\item An input-output analyzer which generates the transformation embedding describing the relationship between input and output strings;
		\item A program generator which produces the function/arguments embeddings over a few steps;
		\item A symbol selector to decode the function/arguments embeddings and generate human readable symbols.
	\end{itemize}
	
	The program generator will run for a few steps, each step it may access the outputs of previous steps, thus enables the model to compose powerful programs using only a few predefined atomic functions. While traditional deep learning models try to generate the output $y$ given $x$ and eventually fit the function $\f$ such that $y=\f(x)$, our model learns to generate the function $\f$ directly and fits the higher-order function $\g$ for which $\f=\g(x,y)$ and $\f$ satisfies $y=\f(x)$. %Our model can be viewed as \emph{learning to learn} or \emph{meta-learning} \cite{lake2016building} and uses few (exactly one) example to generate the desired function.
	\subsection{String Encoder}
	%\begin{figure}[!h]
	%	\centering
	%	\includegraphics[width=1\linewidth]{string_encoder.pdf}
	%	\caption{Diagram of String Encoder}
	%	\label{fig:encoder}
	%\end{figure}
	
	Given a string composed of a sequence of characters $\{i_1,i_2,...,i_L\}$, the string encoder outputs a string embedding and a list of character embeddings. First each character $i_k$ in the sequence is mapped to the 8-dimensional raw embeddings $\bm{e}(i_k)$ via a randomly initialized and trainable embedding matrix. %First the integer-encoded characters $i_k$ in sequence are mapped to the 8-dimensional raw embeddings $\bm{e}(i_k)$ via a randomly initialized and trainable embedding matrix. 
    To better present and attend to a character, the context of each character is fused with the character's raw embedding to build the character embedding. Let $\bm{l}(i_k)$ denote the left context of the character $i_k$ and $\bm{r}(i_k)$ as the right context of $i_k$. $\bm{l}(i_k)$ and $\bm{r}(i_k)$ are respectively calculated using Equation \eqref{eq:c_l} and Equation \eqref{eq:c_r} with $\bm{l}(0)=\bm{r}(L+1)=[0]^C$. In our implementation $\f_l$ and $\f_r$ are the update function of LSTM \cite{Hochreiter1997Long}. 
	\begin{align}
	\label{eq:c_l}\bm{l}(i_k)&=\f_l(\bm{l}(i_{k-1}),\bm{e}(i_k))\\
	\label{eq:c_r}\bm{r}(i_k)&=\f_r(\bm{r}(i_{k+1}),\bm{e}(i_k))
	\end{align}
	The character embedding for each character is the combination of left/right contexts of character $i_k$ and $\bm{e}(i_k)$ itself as shown in Equation \eqref{eq:c_full_max}, where $[\cdot;\cdot]$ means the concatenation of vectors, $\max$ is the element-wise max-pooling. $W_e$ and $\bm{b}_e$ are parameter matrix and vector for building the full character embedding.
	\begin{equation}
	\label{eq:c_full_max}\bm{c}_k=\tanh(W_e[\max(\bm{l}(i_k),\bm{r}(i_k));\bm{e}(i_k)]+\bm{b}_e)
	\end{equation}
$\bm{c}_k$ will be used by the attention mechanism in the program generator. 

We also need a representation which can summarize the whole string, so we can induce the transformation
%transformation embedding 
from the string embeddings of input and output strings. We use multilayer bidirectional LSTM \cite{Graves2005Framewise} to summarize $\bm{c}_k$. The output of forward and backward LSTM at each layer are concatenated and become the input of next layer's forward and backward LSTM. The topmost layer's last hidden states of forward and backward LSTM are merged to generate the string embedding $\bm{s}\in R^S$ through a final fully connected layer. The processing for input and output strings is separated but shares the same neural network architecture and parameters, thus producings  $\bm{c}_{I,1},...\bm{c}_{I,L},\bm{s}_{I}$ and $\bm{c}_{O,1},...\bm{c}_{O,L},\bm{s}_{O}$ for input and output strings, respectively.

	\subsection{Input/output Analyzer}
	The input/output analyzer converts the input and output string embeddings to the transformation embedding, which describes the relationship between the input and output strings. Let $\bm{t}\in R^T$ denote the transformation embedding. $\bm{s}_{I}\in R^S, \bm{s}_{O}\in R^S$ are the input and output string embeddings, respectively. The input/output analyzer can be represented as Equation \eqref{eq:program_gen}. In our implementation, $\f_{IO}$ is just a 2-layer fully connected neural network with activation function $\tanh$.
	\begin{equation}
	\label{eq:program_gen}\bm{t}=\f_{IO}([\bm{s}_{I};\bm{s}_{O}])
	\end{equation}
		\begin{figure}[ht]
		\includegraphics[width=1\linewidth]{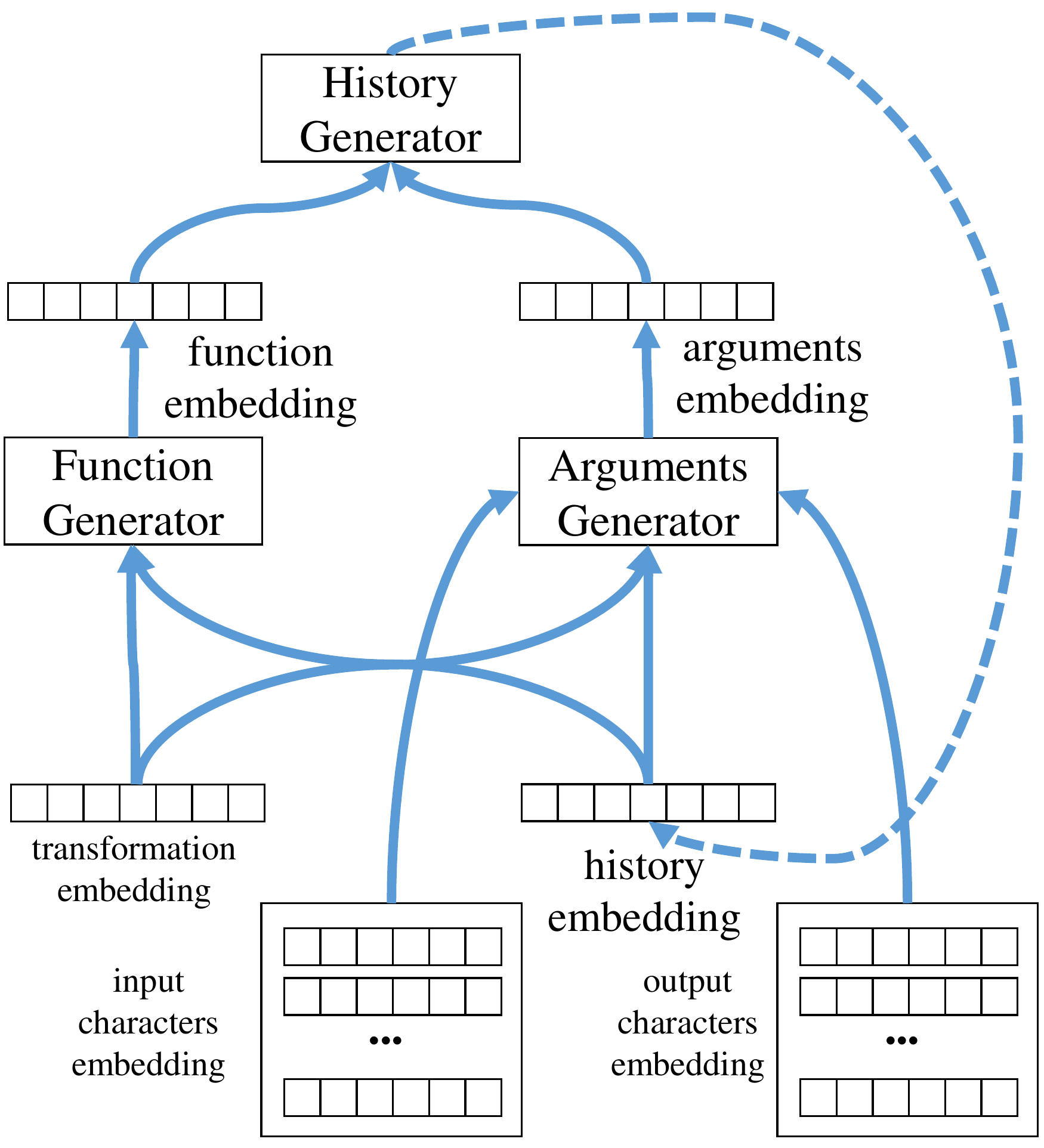}
		\caption{Diagram of the program generator.}
		\label{fig:generator}
 %       \vspace{-12pt}
	\end{figure}
    
	\subsection{Program Generator}
	\begin{table*}[!ht]
		\renewcommand{\arraystretch}{1}
		\centering
		\begin{tabular}{l|l|l|l}
			\hline
			\textbf{Function Name} & \textbf{Arguments}    & \textbf{Return Type} & \textbf{Description}                                            \\
			\hline
			Split        & (String str, Char delim)        & String{[}{]} & Split string using $delim$ as a delimiter, return an array of strings       \\
			Join         & (String{[}{]} strs, Char delim) & String       & Join an array of strings using $delim$ as a delimiter, return a string      \\
			Select       & (String{[}{]} strs, Int index)    & String       & Select an element of a string array, return a string                    \\
			ToLower      & (String str)                    & String       & Turn a string $str$ into lower case                                       \\
			ToUpper      & (String str)                    & String       & Turn a string $str$ into upper case                                       \\
			Concatenate  & (String str or Char c, ...)     & String       & Concatenate a sequence of strings or characters into a string           \\
			GetConstString  & (null)                       & String       & Special function indicating a constant string               \\
			NoFunc       & (null)                          & null         & Special symbol indicating no more function is needed                    \\
			\hline             
		\end{tabular}
		\caption{Atomic functions.}
		\label{table:atom-functions}
	\end{table*}
	The program generator (Figure \ref{fig:generator}) is the core of NPBE. It generates several functions and corresponding arguments' embeddings step by step. The function's embedding at time $t$ is calculated as Equation \eqref{eq:func_embed}, which is a fully connected neural network taking as input the transformation embedding $\bm{t}$ and the execution history of the program generator $\bm{h}_{t-1}\in R^H$ ($\bm{h}_{0}=\bm{t}$).
    %which is a fully connected neural network taking the transformation embedding $\bm{t}$ and the history of running the program generator embedded in $\bm{h}_{t-1}\in R^H$ with $\bm{h}_{0}=\bm{t}$ as input.
    Similarly, the function's arguments embedding at time $t$ is calculated as Equation \eqref{eq:args_raw_embed}. However, the function's arguments are often very complex and hard to predict accurately. So attention mechanism (similar to the one used in neural machine translation models \cite{bahdanau2014neural}) is applied to refine the raw arguments embedding $\bm{a}_{r,t}$ with attention on input and output strings. This is summarized in Equation \eqref{eq:att_args_embed}. Finally, the function embedding $\bm{f}_t$, the refined arguments embedding $\bm{a}_t$, and the previous history embedding $\bm{h}_{t-1}$ are merged into the new history embedding $\bm{h}_t$ as shown in Equation \eqref{eq:history}. $W_h$ and $\bm{b}_h$ are parameter matrix and vector for generating the new history, respectively.
    %the process is summarized as Equation \eqref{eq:att_args_embed}. Finally, the function embedding, the refined arguments' embedding and the previous history embedding are merged into the new history embedding as shown in Equation \eqref{eq:history}, where $W_h$ and $\bm{b}_h$ are parameter matrix and vector for generating new history, respectively.
	\begin{align}
	\label{eq:func_embed}\bm{f}_t&=\f_{func}(\bm{t},\bm{h}_{t-1})\\
	\label{eq:args_raw_embed}\bm{a}_{r,t}&=\f_{args}(\bm{t},\bm{h}_{t-1})\\
	\label{eq:att_args_embed}\bm{a}_{t}&=\f_{att}(\bm{a}_{r,t},\bm{c}_{I,1},...\bm{c}_{I,L},\bm{c}_{O,1},...\bm{c}_{O,L})\\
	\label{eq:history}\bm{h}_{t}&=\tanh(W_{h}[\bm{f}_t;\bm{a}_{t};\bm{h}_{t-1}]+\bm{b}_h)
	\end{align}
	Functions $\f_{func}$ and $\f_{args}$ can be multilayer fully connected neural networks, and in our experiment we just use one layer neural network. The function $\f_{att}$ in Equation \eqref{eq:att_args_embed} is implemented by attending to the input and output strings as follows: \\
	\begin{align}
	u_{I,i}^t&=\bm{v}_{I}^T\tanh(W_{1}\bm{c}_{I,i}+W_{2}\bm{a}_{r,t})&i\in(1,...,L)\\
	a_{I,i}^t&=\softmax(u_{I,i}^t)&i\in(1,...,L)\\
	\bm{a}_{I,t}&=\sum_{i=1}^{L}a_{I,i}^t\bm{c}_{I,i}&\\
	u_{O,i}^t&=\bm{v}_{O}^T\tanh(W_{3}\bm{c}_{O,i}+W_{4}\bm{a}_{r,t})&i\in(1,...,L)\\
	a_{O,i}^t&=\softmax(u_{O,i}^t)&i\in(1,...,L)\\
	\bm{a}_{O,t}&=\sum_{i=1}^{L}a_{O,i}^t\bm{c}_{O,i}&
	%\bm{a}_{t}&\hskip=\tanh(W_{att}[\bm{a}_{I,t};\bm{a}_{R,t};\bm{a}_{O,t}]+\bm{b}_{att})
	\end{align}
	$W_1, W_2, \bm{v}_I$ and $W_3, W_4, \bm{v}_O$ are parameters for attending to the input and output strings, respectively. Note that the attention architecture for input and output strings are the same but with different parameters. The final arguments embedding is generated by combining attention over input, output and the raw arguments embedding:
	\begin{equation}
	\bm{a}_{t}=\tanh(W_{att}[\bm{a}_{I,t};\bm{a}_{r,t};\bm{a}_{O,t}]+\bm{b}_{att})
	\end{equation}
	where $W_{att}$ and $\bm{b}_{att}$ are parameters for combining.
	\subsection{Symbol Selector}
	The symbol selector uses the function embedding $\bm{f}_t$ and the arguments embedding $\bm{a}_t$    generated by the program generator to select a proper function and the corresponding arguments of that function. The probability distribution $\bm{\alpha}_{func,t}\in [0,1]^{P}$ over $P$ atomic functions is produced by Equation \eqref{eq:func_symbol}, where $U_f\in R^{P\times F}$ is the matrix storing the representations of the atomic functions. The arguments embedding $\bm{a}_t$ (representing the summary of the arguments) is decoded by a RNN \cite{Cho2014Learning}, which is conditioned on its previous output and $\bm{a}_t$, as shown in Equation \eqref{eq:args_rnn}. $\bm{s}_{t,0},...\bm{s}_{t,i-1}$ are hidden states of LSTM with $\bm{s}_{t,0}=[0]^H$. In this way, the sequence of arguments is generated by the RNN. The probability distribution of the $i$-th argument at time $t$ over $Q$ possible arguments, $\bm{\alpha}_{arg,i,t}\in [0,1]^{Q}$, is produced using Equation \eqref{eq:args_symbol}, where  $U_a\in R^{Q\times A}$ is the matrix storing the representations of possible arguments.
%The arguments embedding $\bm{a}_t$ is decoded by an RNN which has previous output and as input and a peak over initial $\bm{a}_t$ like Equation \eqref{eq:args_rnn}. $\bm{s}_{t,0},...\bm{s}_{t,i-1}$ are hidden states of LSTM with $\bm{s}_{t,0}=[0]^H$. Thus we get the embeddings of each argument.
%Then the probability distribution of the $i$-th argument at time $t$ $\bm{\alpha}_{arg,i,t}\in [0,1]^{Q}$ over $Q$ possible arguments is produced in a similar way as function by Equation \eqref{eq:args_symbol} where  $U_a\in R^{Q\times A}$ is the matrix storing the representations of possible arguments.
	\begin{align}
	\label{eq:func_symbol}\bm{\alpha}_{func,t}&=\softmax(U_f\bm{f}_t)\\
	\label{eq:args_rnn}\bm{a}_{t,i}&=\LSTM(\bm{s}_{t,i-1},\bm{a}_{t,i-1},\bm{a}_t)\\
	\label{eq:args_symbol}\bm{\alpha}_{arg,i,t}&=\softmax(U_a\bm{a}_{t,i})
	\end{align}
	\subsection{Training}
	We train the NPBE model end-to-end using input-output string pairs $\{\mathscr{X}_i,\mathscr{Y}_i\}$ as well as the programs represented as a sequence of functions and corresponding arguments. Each program $\mathscr{P}_{i}:\{\f^{1}_{i},\bm{a}^{1}_{i},...,\f^{T}_{i},\bm{a}^{T}_{i}\}$ can transform $\mathscr{X}_i$ to $\mathscr{Y}_i$, where $i$ means the $i$'th training example. For every input-output pair we can generate a sequence of $T$ functions and every function has an argument list $\bm{a}\in A^M$, where $A$ is the set of possible arguments, $M$ is the maximum number of arguments one function can take. In our implementation, $T=5,M=5$.
%	We train the NPBE model end-to-end using input-output string pairs $\{\mathscr{X}_i,\mathscr{Y}_i\}$ as input, and the program represented as a sequence of functions and corresponding arguments $\mathscr{P}_{i}:\{\f^{1}_{i},\bm{a}^{1}_{i},...,\f^{T}_{i},\bm{a}^{T}_{i}\}$ which can transform $\mathscr{X}_i$ to $\mathscr{Y}_i$ as training target, where $i$ means the i'th training example. For every input-output pair we can generate a sequence of $T$ functions and every function has an arguments list $\bm{a}\in A^M$, where $A$ is the set of possible arguments, $M$ is the maximum number of arguments one function can take. In our implementation, $T=5,M=5$.
	
	The training is conducted by directly maximizing the log-likelihood of the correct program $\mathscr{P}$ given $\{\mathscr{X},\mathscr{Y}\}$:
	\begin{equation}
	\theta^*=\argmax_{\theta}\sum_{(\mathscr{X},\mathscr{Y},\mathscr{P})}\log P(\mathscr{P}|\mathscr{X},\mathscr{Y};\theta)
	\end{equation}
	where $\theta$ is the parameters of our model. Random Gaussian noise \cite{neelakantan2015adding} is injected into the transformation embedding and the arguments embedding to improve the generalization ability and stability of NPBE.
	
	\begin{table}[t]
		\renewcommand{\arraystretch}{1}
		\centering
		\begin{tabular}{p{0.3\linewidth}|p{0.6\linewidth}}
			\hline
			\textbf{Constant Type} & \textbf{Constant} \\
			\hline                                        
			Delimiter      & ``(space)'', ``(newline)'', ``,'', ``.'', ``\textbackslash'', ``@'', ``:'', ``;'', ``\_'', ``='', ``-'', ``/'' \\
			Integer        & 0, 1, 2, 3, -1, -2, -3                                            \\
			Special Symbol & $x, o1, o2, o3, o4, NoArg$ \\
			\hline                               
		\end{tabular}
		\caption{Constant symbols of our model.}
		\label{table:atom-constants}
	\end{table}

	\begin{table*}[ht]
		\renewcommand{\arraystretch}{1}
		\centering
		\begin{tabular}{l|c|c|c|c|c|c|c}
			\hline
			\textbf{Input-output Pair} & \textbf{t} & \textbf{Function} & \textbf{Argument 1} & \textbf{Argument 2} & \textbf{Argument 3} & \textbf{Argument 4} & \textbf{Argument 5}\\
			\hline
			Input string: & 1 & GetConstString & --- & --- & --- & --- & --- \\
			john@company.com & 2 & GetConstString & --- & --- & --- & --- & --- \\
			Output string: & 3 & Split & x & ``@'' & --- & --- & --- \\
			Hello john, have fun!& 4 & Select & o3 & 0 & --- & --- & --- \\
			& 5 & Concatenate & o1 & o4  & o2 & --- & --- \\
			%\hline
			%\multicolumn{2}{c|}{\textit{Composed Program:}} &
			%\multicolumn{6}{l}{Concatenate(GetConstString(), Select(Split(x, ``@''), 0), GetConstString())} \\
			\hline
			Input string: & 1 & Split & x & ``/'' & --- & --- & --- \\
			17/apr/2016 & 2 & Select & o1 & 0 & --- & --- & --- \\
			Output string: & 3 & Select & o1 & 1 & --- & --- & --- \\
			APR-17 & 4 & ToUpper & o3 & ---  & --- & --- & --- \\
			& 5 & Concatenate & o4 & ``-''  & o2 & --- & --- \\
			%\hline
			%\multicolumn{2}{c|}{\textit{Composed Program:}} & \multicolumn{6}{l}{\textit{Concatenate(ToUpper(Select(Split(x, ``/''),  1)), ``-'', Select(Split(x, ``/''), 0))}} \\
			\hline
% 			Input string: & 1 & Split & x & ``/'' & --- & --- & --- \\
% 			1999/08/17 & 2 & Select & o1 & 0 & --- & --- & --- \\
% 			Output string: & 3 & Select & o1 & 1 & --- & --- & --- \\
% 			08-17,1999 & 4 & Select & o1 & 2 & --- & --- & --- \\
% 			& 5 & Concatenate & o3 & ``-'' & o4 & ``,'' & o2 \\
% 			\hline
			
			Input string: & 1 & Split & x & ``/'' & --- & --- & --- \\
			/home/foo/file.cpp & 2 & Select & o1 & -1 & --- & --- & --- \\
			Output string: & 3 & Split & o2 & ``.'' & --- & --- & --- \\
			file & 4 & Select & o3 & 0 & --- & --- & --- \\
			%\hline
			%\multicolumn{2}{c|}{\textit{Composed Program:}} & \multicolumn{6}{l}{Select(Split(Select(Split(x, ``/''), 1), ``-''), 1)} \\
			\hline
		\end{tabular}
		\caption{Examples of input-output pairs for NPBE. The corresponding programs (from top to bottom) are:
			1) \textit{Concatenate(``Hello '', Select(Split(x, ``@''), 0), ``, have fun!'')}; 2) \textit{Concatenate(ToUpper(Select(Split(x, ``/''),  1)), ``-'', Select(Split(x, ``/''), 0))}; and 3) \textit{Select(Split(Select(Split(x, ``/''), -1), ``.''), 0)}. Note that \textit{GetConstString} is a placeholder, and the first one is evaluated to ``Hello '', the second one is evaluated to ``, have fun!''.}
		\label{table:pattern-io-examples}
	\end{table*}

	\section{Experiments}

	\subsection{Experimental Design}
	The NPBE model is required to induce a program consisting of a sequence of functions based on only one input-output string pair. In this section, we describe our evaluation of the NPBE model.
	%To achieve this task, the model should be able to 1) analyze the input-output string pair and find their pattern; 2) learn to use each atom functions correctly; 3) compose a complex program using atomic functions; 4) even generalize to unseen arguments combinations. 
	In our experiments, we define 7 basic string manipulation functions and 1 null function as the atomic functions (Table \ref{table:atom-functions}). For simplicity, each program is allowed 
	to be composed by 5 functions at most. We define a set of constant symbols (Table \ref{table:atom-constants}) from which our model can choose as arguments. The constant symbols include integers and delimiters. The integers are used by the $Select$ function as index to an array of strings. The negative integers are used to access array elements from the tail. The design of integer symbols supports the access to an array of at most 7 elements. The delimiters are used by $Split$ and $Join$ to split a string or join an array of strings by a delimiter. We also define some special symbols. For example, the symbol $x$ refers to the input string, $o1, o2, o3$ and $o4$ are used to refer to the output of the first, second, third and fourth operation, respectively. The $NoArg$ symbol indicates that no arguments is expected at the current position. Note that although in our experiments, we give some constraints to the functions and arguments in a program, our model can be easily extended to support new functions and arguments.

    To obtain training data, we first generate programs at various levels of complexity according to 45 predefined tasks.
    A task is a sequence of function in a specific order, but the arguments of each function are not fixed. %There may exist many possible programs given a task. 
	The reason for defining tasks is that we want the program generated by our model being syntactically correct and meaningful. The 45 tasks range from simple ones such as the concatenation of input to some constant string, to more complex ones comprising \textit{Split, Join, Select, ToUpper} and \textit{Concatenate} functions. For example, the task of \textit{Split, Join} is to first split the input string by a delimiter, then join the split strings array using another delimiter. A program derived from this task could be \textit{Join(Split(x,``/"), ``:")}, which splits the input string $x$ according to the delimiter ``/" and then joins the resulting substrings using the delimeter ``:". The average number of functions for accomplishing a task is 3.5.
    %The reason for defining tasks is that we want the program generated by our model being syntactically correct and meaningful. The 45 tasks range from simple programs such as the concatenation of input to some constant string, to more complex programs comprising \textit{Split, Join, Select, ToUpper} and \textit{Concatenate} functions. For example, the task of \textit{Split, Join} is to first split the input string by a delimiter, then join the split strings array using another delimiter. The average number of functions for accomplishing a task is 3.5.
    %The average steps of operation to accomplish a task is 3.5.

%For all the tasks, we generate a total number of 90,000 programs. We choose about 69,000 programs to train our model and keep the remaining programs unseen by the model for testing. 
For all the tasks, we generate a total number of around 69,000 programs for training. For each program $\mathscr{P}_i$ we generate a random input string $\mathscr{X}_i$, which should be an valid input to the program $\mathscr{P}_i$. Next, we apply the program $\mathscr{P}_i$ on $\mathscr{X}_i$ by actually running the Python program implementing $\mathscr{P}_i$ and get the output string $\mathscr{Y}_i$. We constrain $\mathscr{X}_i$ and $\mathscr{Y}_i$ to be at most 62 characters long. After that, we use $\{\mathscr{X}_i,\mathscr{Y}_i\}$ as the input data to our model, and $\mathscr{P}_i$ as the training target. The program is generated in such a way that if there are multiple programs that can result in the same $\mathscr{Y}_i$ given $\mathscr{X}_i$, only one specific program is chosen. Therefore, there is no ambiguity for the model to predict the desired program. Given the program $\mathscr{P}_i$, the input string $\mathscr{X}_i$ is always generated dynamically and randomly to decrease overfitting. %The evaluation of NPBE is conducted on randomly generated input-output pairs too. 
%	For these tasks, the total amount of all possible programs is about 90,000. We choose about 69,000 programs to train our model and keep the remaining programs unseen by the model for testing. Then according to the program $\mathscr{P}_i$ we generate a random input string $\mathscr{X}_i$ which must be valid as the input of program $\mathscr{P}_i$. Next, we apply the program $\mathscr{P}_i$ on $\mathscr{X}_i$ by actually running the python program represented by $\mathscr{P}_i$ and get $\mathscr{Y}_i$. We constrain $\mathscr{X}_i$ and $\mathscr{Y}_i$ to be at most 62 characters long. After that, we use $\{\mathscr{X}_i,\mathscr{Y}_i\}$ as the input to our model, and $\mathscr{P}_i$ as the training target. The program is generated in such a way that if there are multiple programs which can result in the same $\mathscr{Y}_i$ given $\mathscr{X}_i$, only one specific program is chosen. Therefore, there is no ambiguity for the model to predict the desired program. Given the program $\mathscr{P}_i$, the input string $\mathscr{X}_i$ is always generated dynamically and randomly to decrease overfitting. %The evaluation of NPBE is conducted on randomly generated input-output pairs too. 
	Table \ref{table:pattern-io-examples} gives some concrete input-output examples for our model.

	%We define 64,000 examples of training as one epoch, but actually since all data are generated dynamically the possibility of seeing one example twice is very low.
	To train NPBE, we choose RMSProp \cite{tieleman2012lecture} as the optimizer and set the mini-batch size to 200. We set the dimensionality of the transformation embedding $\bm{t}$ and the history embedding $\bm{h}$ to 256, the function embedding $\bm{f}$ to 16, and the arguments embedding $\bm{a}$ to 64.
	The actual training process relies on an adaptive curriculum ~\cite{reed2015neural} in which the frequency of one specific task being trained is proportional to its error rates over test. Every 10 epochs we estimate the prediction errors. We use softmax with adequate temperature over error rates to sample the frequency of each task that will be trained during the next 10 epochs. So tasks with the higher error rates will be sampled more frequently than tasks with the lower error rates during the next 10 epochs.
	
The evaluation of NPBE is conducted to answer the following research questions:

	\begin{table*}[!ht]
		\centering
		\begin{tabular}{l|c|c|c|c|c}
			\hline
			\textbf{Task} & \textbf{LSTM} & \textbf{LSTM-A} & \textbf{NPBE-Top1} & \textbf{NPBE-Top3} & \textbf{NPBE-Top5}\\
			\hline
			Case Change & 100.0\% & 100.0\% & 100.0\% & 100.0\% & 100.0\% \\
			Duplicate Input String & 9.4\% & 9.8\% & 99.1\% & 99.1\% & 99.1\% \\
			Case Change and Concatenate with Input String & 9.4\% & 9.3\% & 99.9\% & 99.9\% & 99.9\%  \\
			Concatenate with Constant & 83.1\% & 67.5\% & 88.4\% & 88.4\% & 88.4\%  \\
			Split, Select, Case Change  & 8.7\% & 12.8\% & 92.3\% & 96.4\% & 97.3\% \\
			Split, Join, Select, Case Change, Concatenate  & 0.1\% & 0.1\% & 92.4\% & 96.6\% & 97.8\% \\
			GetConstString, Split, Select, Case Change, Concatenate  & 2.9\% & 5.2\% & 81.2\% & 85.6\% & 86.7\% \\

			GetConstString$\times2$, Split, Select, Concatenate & 0.1\% & 0.1\% & 24.7\% & 54.4\% & 68.1\% \\

			Split, Select, Select, Concatenate & 0.1\% & 0.1\% & 9.8\% & 46.4\% & 82.0\% \\
			
			Split, Select, Select, Case Change, Concatenate & 0.1\% & 0.2\% & 35.8\% & 73.2\% & 94.0\% \\
			Split, Select, Split, Select & 0.1\% & 0.1\% & 31.2\% & 64.3\% & 72.2\% \\
			\hline
			Average over All 45 Tasks & 20.4\% & 22.0\% & 74.1\% & 85.8\% & 91.0\% \\
			\hline                               
		\end{tabular}
		\vspace{-6pt}
%		\caption{Results on predicting programs.%LSTM-A means LSTM encoder-decoder with attention. Note that we only consider alternative arguments for $Select$, so tasks without the use of $Select$ don't have top3 and top5 accuracy.
		\caption{Results of generating programs.
		}
		\vspace{-6pt}

		\label{table:results}
	\end{table*}
	\begin{table}[!ht]
		\renewcommand{\arraystretch}{1}
		\centering
		\begin{tabular}{l|c|c}
			\hline
			\textbf{Task} & \textbf{Seen} & \textbf{Unseen}\\
			\hline
			Split, Join & 93.6\% & 93.9\% \\
			GetConstString, Split, Join, Concatenate & 91.4\% & 91.5\% \\
			Split, Join, Concatenate & 95.0\% & 94.9\% \\
			Split, Join, Select, Concatenate & 90.8\% & 89.5\% \\
			Split, Select, Select, Concatenate & 82.0\% & 82.1\% \\
			\hline
			Average over 19 Tasks & 87.6\% & 87.4\% \\
			\hline                              
		\end{tabular}
        \vspace{-6pt}
%		\caption{Results on seen and unseen program 
		\caption{Results with seen and unseen program 
arguments.}
		\vspace{-6pt}
		\label{table:never_seen_trans}

	\end{table}

	\subsubsection{RQ1: What is the accuracy of NPBE in generating programs?}
	In this RQ, we use randomly generated input-output strings to evaluate the accuracy of NPBE in generating programs. For example, given a random input-output pair: \textit{25/11/16} and \textit{25:11:16} (which does not appear in the training data), we would like to test if the correct program \textit{Join(Split(x, ``/''), ``:'')} can be still generated. To answer this RQ, we generate random input-output strings for each task 1000 times and apply the trained NPBE model. A program produced by NPBE is regarded correct only if the model predicts all the five functions (if less than five, padded with the $NoFunc$ symbol) and all the arguments of the functions (also padded with the $NoArg$ symbol) correctly (thus a total of $5+5\times5=30$ positions).   
    We also compare our model with the RNN encoder-decoder models \cite{Cho2014Learning} implemented using LSTM and LSTM with attention mechanism, which all have similar total number of parameters to our model.

	\subsubsection{RQ2: Can NPBE generate programs with previously unseen arguments?}
 	In this RQ, we test the generalization ability of NPBE. We evaluate our model using programs whose argument settings do not appear in the training set. For example, if the program \textit{Join(Split(x, ``/''), ``:'')} appears in the training set, we would like to know if NPBE can work for the program \textit{Join(Split(x, ``@''), ``-'')} , which does not appear in the training set. To answer this RQ, we design a test set consisting of around 19,000 programs with previously unseen arguments. The experiment is conducted on 19 selected tasks that have complex argument combinations (for simple tasks there are few arguments to choose so we skip them). 
%	In this RQ, we test the generalization ability of NPBE. We evaluate our model with program arguments settings that do not appear in the training set. For example, if the program \textit{Select(Split(x, ``@''), 0)} appears in the training set, we would like to know if NPBE can work for the program \textit{Select(Split(x, ``/''), 1)}, which does not appear in the training set.
	
%	\subsection{RQ3: What's the advantage and disadvantage of NPBE compared to traditional PBE methods?}
%	In this RQ, we try to analyze the advantage and disadvantage of NPBE compared to traditional PBE methods and give directions of future researches of our model.

% 	\subsection{RQ3: What is the accuracy of NPBE in generating known programs with longer strings?}
% 	In this RQ, we evaluate the performance of some tasks by providing input-output strings which are longer than the maximum input-output strings in the training set.
	
	\subsection{Experimental Results}

	\subsubsection{RQ1: What is the accuracy of NPBE in generating programs?}
	
	%Our evaluation shows that NPBE can successfully generate programs based on only one input-output pair.
	Table \ref{table:results} gives the evaluation results of NPBE on predicting programs. %with arguments which have occurred exactly in the training data.
     The average Top1 accuracy achieved by NPBE is 74.1\%, which means that for 74.1\% of input-output pairs in test, NPBE successfully generates the corresponding program. We found that the model prediction errors most likely to occur on the integer argument of $Select$ because neural networks are not good at counting. So we also let the model to give 3 or 5 predictions when it tries to predict the integer arguments of $Select$. The average Top3 and Top5 accuracy results are 85.8\% and 91.0\%, which means that for 85.8\% and 91.0\% of input-output pairs in test, NPBE successfully returns the corresponding program within the top 3 and top 5 results respectively. The results show that the NPBE model can generate correct programs for most tasks. 

%	The result is regarded as correct only if the model predicts the five functions (if less than five, padded with $NoFunc$ symbol) and every arguments of each function (also padded with $NoArg$ symbol, thus a total of $5+5\times5=30$ positions) right. The accuracy of each task is obtained by generating completely random arguments and corresponding input-output strings of that task for 1000 times and testing on those examples. We found that the model prediction errors most likely to occur on the integer argument of $Select$ because neural networks are not good at counting. So we set the model to give 3 or 5 predictions when it tries to predict the integer arguments of $Select$. The average Top1 accuracy achieved by NPBE is 74.1\%, which means that for 74.1\% of input-output pairs in test, NPBE successfully generates the corresponding program. The average Top3 and Top5 accuracy results are 85.8\% and 91.0\%, which means that for 85.8\% and 91.0\% of input-output pairs in test, NPBE successfully returns the corresponding program within the top 3 and top 5 results respectively. The results show that the NPBE model can generate correct programs for most tasks. 
	
    As an example, given the input string ``17/apr/2016'' and output string ``APR-17'', our model needs to induce a program comprising \textit{Split, Select, Select, Case Change, Concatenate}. For this task, % (can be harder than this concrete example because the input string can be split into at most 7 substrings), 
our model gives completely correct program in 35.8\% cases. If we allow the model to give 3 or 5 predictions for the integer argument of $Select$, the accuracy is increased to 73.2\% or 94.0\%. %Note that the accuracies are got on test examples which are harder than this concrete example because the input string can be split into at most 7 substrings.
	
	The results about LSTM and LSTM with attention mechanism (denoted to as LSTM-A) are also shown in Table \ref{table:results}. Note that the Top1, Top3 and Top5 accuracy for LSTM and LSTM-A are almost the same, so only the Top1 accuracy is reported. We found that the ordinary encoder-decoder model can solve the simplest tasks but cannot tackle harder tasks. The results show that NPBE significantly outperforms the ordinary encoder-decoder model.
	
% 	In summary, our evaluation results confirm that NPBE can successfully generate programs based on input and output examples, and outperforms the RNN encoder-decoder based approaches.

	\subsubsection{RQ2: Can NPBE generate programs with previously unseen arguments?}
	
	We test the generalization ability of NPBE on the programs with previously unseen argument settings. %The experiment is conducted on 19 selected tasks that have complex argument combinations (for simple tasks there are few arguments to choose). 
    The average Top5 accuracy results are given in Table \ref{table:never_seen_trans}. The results shows that for seen and unseen argument settings the accuracies achieved by our model have no big difference. For example, the task of \textit{Split, Join} first splits an input string by a delimiter (such as ``/'') and then concatenates the resulting substrings by the other delimiter (such as ``:''). This task achieves 93.6\% accuracy on the training set. For different arguments (e.g., first split the input string by ``@'' then join the resulting substrings by ``-'') that do not exist in the training set, NPBE can still get 93.9\% accuracy. The results show that NPBE can be generalized to unseen program arguments without over-fitting to particular argument combinations.  
 	
 	%That means NPBE can use each argument freely without overfitting to particular argument combinations.  %the model can be generalized to unseen program arguments.

\subsection{Discussions and Future Work} 
%In the experiments, we found that tasks with Case Change usually have higher accuracy than similar tasks without Case Change (such as the tasks \textit{Split, Select, Select, Concatenate} and \textit{Split, Select, Select, Case Change, Concatenate}). This indicates that our model can recognize Case Change as a feature or a cue \cite{menon2013machine} to help predict correct program. 
The intention behind NPBE is to make the model learn related features from input-output strings automatically and use the learned features to induce correct programs. The purpose of this paper is not to directly compete with the existing PBE systems. Instead, we show that the use of DNN can recognize features in string transformations and can learn accurate programs through input-output pairs. %The idea of this paper can also be applied in conjunction with prior systems to speed up the search for possible programs. 

Currently, NPBE cannot be generalized to completely unseen tasks (such as \textit{Split, Join, Join, Concatenate}) that never appeared in the training set. In our future work, we will try to build the model that really ``understands'' the meaning of atomic functions to make it possible to generalize to the unseen tasks. %Also, we plan to improve the model to take more than one input-output pairs to increase the prediction accuracy.

	\section{Conclusion}
%	\blfootnote{Source code and demo can be downloaded at: https://sites.google.com/site/npbepaper/}
	In this paper, we propose NPBE, a Programming by Example (PBE) model based on DNN. NPBE can induce string manipulation programs based on simple input-output pairs by inferring a composition of functions and corresponding arguments. We have shown that the novel use of DNN can be successfully applied to develop Programming By Example systems. Our work also explores the way of learning higher-order functions in deep learning, and is one step towards teaching DNN to generate computer programs.
% The source code of NPBE and a demo can be downloaded at: https://sites.google.com/site/npbepaper.
	
%\section{Acknowledgements}
  %  This work was done during an internship at Microsoft Research Asia. We would like to thank anonymous reviewers for their careful reviews and pertinent comments.
	\bibliographystyle{aaai}
	\balance
	\bibliography{reference}

\end{document}